\documentclass[a4paper,fleqn]{cas-dc}
\usepackage{soul}
\usepackage[numbers]{natbib}
\usepackage{algorithm}
\usepackage{algpseudocode}
\usepackage{pifont}
\newcommand{\cmark}{\ding{51}}
\newcommand{\xmark}{\ding{55}}
\usepackage{bm}

\usepackage{tabularx}
\definecolor{brightgreen}{rgb}{0.0, 0.5, 0.0}
\def\tsc#1{\csdef{#1}{\textsc{\lowercase{#1}}\xspace}}
\tsc{WGM}
\tsc{QE}
\tsc{EP}
\tsc{PMS}
\tsc{BEC}
\tsc{DE}

\begin{document}
\let\WriteBookmarks\relax
\def\floatpagepagefraction{1}
\def\textpagefraction{.001}

\shorttitle{}
\shortauthors{C. Thai, M. X. Trang et~al.}

\title [mode = title]{Enhancing Rotated Object Detection via Anisotropic Gaussian Bounding Box and Bhattacharyya Distance}                      


%

\author[1, 2]{Chien Thai}[type=editor, auid=000, bioid=1, orcid=0000-0002-5098-6862]
\ead{chientv@phenikaa-x.com}

\author[1]{Mai Xuan Trang}[orcid=0000-0002-9666-0198]
\cormark[1]
\ead{trang.maixuan@phenikaa-uni.edu.vn}

\author[3]{Huong Ninh}
\ead{huongnt382@viettel.com.vn}

\author[2]{Hoang Hiep Ly}

\author[2]{Anh Son Le}


\affiliation[1]{organization={Faculty of Computer Science},
    addressline={Phenikaa University}, 
    city={Hanoi},
    postcode={12116}, 
    country={Vietnam}}
\affiliation[2]{origanization={Phenikaa-X Joint Stock Company,},
              addressline={Phenikaa Group},
              city={Hanoi},
              country={Vietnam}}

\affiliation[3]{organization={Optoelectronics Center},
    addressline={Viettel Aerospace Institute, Viettel Group}, 
    city={Hanoi},
    country={Vietnam}}
\cortext[cor1]{Corresponding author}

\begin{abstract}
Detecting rotated objects accurately and efficiently is a significant challenge in computer vision, particularly in applications such as aerial imagery, remote sensing, and autonomous driving. Although traditional object detection frameworks are effective for axis-aligned objects, they often underperform in scenarios involving rotated objects due to their limitations in capturing orientation variations. This paper introduces an improved loss function aimed at enhancing detection accuracy and robustness by leveraging the Gaussian bounding box representation and Bhattacharyya distance. In addition, we advocate for the use of an anisotropic Gaussian representation to address the issues associated with isotropic variance in square-like objects. Our proposed method addresses these challenges by incorporating a rotation-invariant loss function that effectively captures the geometric properties of rotated objects. We integrate this proposed loss function into state-of-the-art deep learning-based rotated object detection detectors, and extensive experiments demonstrated significant improvements in mean Average Precision metrics compared to existing methods. The results highlight the potential of our approach to establish new benchmark in rotated object detection, with implications for a wide range of applications requiring precise and reliable object localization irrespective of orientation.
\end{abstract}



\begin{keywords}
Rotated Object Detection \sep Bounding Box Regression \sep Gaussian Distribution \sep Bhattacharyya Distance
\end{keywords}

\maketitle

\section{Introduction}
Rotated Object Detection, also known as Oriented Object Detection, is a crucial area in computer vision and machine learning that focuses on recognizing and localizing objects within an image regardless of their orientation. Unlike traditional object detection, which typically assumes objects are aligned with the image axes, rotated object detection addresses the challenge of detecting objects that appear at arbitrary angles. This capability is essential for applications where objects may not follow a standard orientation due to camera angles, object movements or natural positions. Rotated object detection is now widely used in a variety of industrial applications, enhancing the versatility and accuracy of automated systems including remote sensing \cite{wen2023comprehensive}, autonomous driving \cite{zhu2023understanding}, scene text detection \cite{he2017deep}, and aerial surveillance \cite{ding2021object}. Therefore, this kind of research has gained much interest in recent years.

\begin{figure}[htbp]
    \centering
    \includegraphics[width=1.0\linewidth]{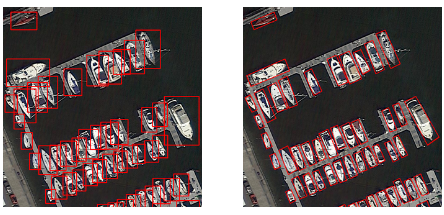}
    \caption{Comparison of Horizontal Bounding Boxes ($x, y, w, h$) and Oriented Bounding Boxes ($x, y, w, h, \theta$)}
    \label{fig:hbb_obb}
\end{figure}

Different from traditional object detection problem, which utilizes a horizontal bounding box (HBB) containing the object center ($x, y$) and size ($w, h$) to represent the location of the object, an oriented bounding box (HBB) uses additional orientation parameter $\theta$ to provide a more accurate representation of object boundaries, thus reducing the overlap with background and improving detection precision. The illustrations of OBB and HBB are shown in Figure \ref{fig:hbb_obb}.

One of the critical components influencing the performance of these systems is the loss function, which plays a pivotal role in guiding the learning process during model training. Conventional loss functions, primarily designed for axis-aligned boxes, often struggle to account for the geometric complexities associated with rotated bounding boxes. Recent rotated object detection frameworks aim to bridge the gap between traditional loss function formulations and the requirements of rotated object detection by introducing innovative modifications that improve both localization accuracy and convergence stability. In horizontal object detection, the Intersection over Union (IoU) is a commonly used metric to measure the overlap between the predicted bounding box and the ground truth bounding box. However, when dealing with rotating object detection problems, directly using IoU loss presents several challenges and limitations, including complex calculation, and non-differentiable in various regions of input space. To address these challenges, numerous works have proposed novel regression losses that approximate the rotating IoU loss function by converting the rotated bounding box to Gaussian representation and utilizing distances between two multivariate Gaussian distribution to quantify the similarity between two bounding boxes \cite{yang2021rethinking, yang2021learning, yang2022kfiou, llerena2021gaussian}. Although Gaussian distribution is an effective methodology to present oriented bounding boxes, it is not an optimal solution for depicting square-like bounding boxes. Specifically, two square-like objects with the similar center and size but different orientations can be represented by a single Gaussian distribution, leading to inaccurate angle prediction. 

In this work, we propose a novel representation for square-like bounding boxes by anisotropically scaling the Gaussian distribution. Additionally, we introduce a new loss function that incorporates modifications to the Bhattacharyya distance \cite{bhattacharyya1943measure} between two multivariate Gaussian distributions, ensuring consistency with the Intersection over Union (IoU) metric. This approach enhances the alignment between distance measures and performance evaluation criteria. The key contributions of this work are summarized as follows:

\begin{itemize}
    \item We demonstrate that the original Gaussian distribution is inadequate for accurately representing square-like objects, as observed in our prediction results. To address this, we propose a novel representation for square-like bounding boxes by introducing anisotropic scaling of the Gaussian distribution.
    \item We explore the use of the Bhattacharyya Distance \cite{bhattacharyya1943measure} for computing the overlap between two rotated bounding boxes and present modifications to align this approach with the IoU loss, enhancing its effectiveness for rotated object detection.
    \item We evaluated the proposed method through extensive experiments on two large-scale datasets for oriented object detection: DOTA \cite{xia2018dota} and HRSC2016 \cite{liu2017high}. Our results indicate that, when integrated into a state-of-the-art deep learning framework, the proposed loss function significantly improve mean Average Precision (mAP) metrics in comparison existing approaches.
\end{itemize}

The paper is organized as follows: Section \ref{sec:related_works} reviews related works on both horizontal and rotated object detection problems. Section \ref{sec:proposed_method} details the proposed method. Section \ref{sec:resutls} presents the datasets, experimental setup, and experimental results. Finally, Section \ref{sec:conclusion} discusses the conclusion and outlines potential future work.

\section{Related Works}
\label{sec:related_works}
\subsection{Backbone Architectures}
Backbone networks are crucial components in modern image processing models, especially in tasks such as image classification, object detection, segmentation, and other vision-related problems. These networks are typically pre-trained on large datasets like ImageNet \cite{deng2009imagenet}, and their learned feature representations can be transferred to other tasks. Marking a breakthrough in image classification, AlexNet \cite{krizhevsky2012imagenet} introduced deep and more efficient architectures using GPUs and ReLU activations. To train very deep networks, ResNet \cite{he2016deep} introduced the concept of residual learning, which allows deeper networks to be trained more effectively. This deep architecture helps the model learn more complex and abstract representations of the input data. Deep Nearest Centroids (DNC) \cite{wang2022visual} proposes a case-based reasoning approach that simplifies classification by using class sub-centroids for proximity-based decisions, making the model flexible, explainable, and easily transferable across tasks with minimal learnable parameters. Challenging CNN-based backbones, Vision Transformers \cite{dosovitskiy2020image} process images as sequences of patches and have shown excellent performance in various tasks.

\subsection{Horizontal Object Detection}
\textbf{Object Detection Framework}: Horizontal object detection in computer vision involves identifying and localizing objects aligned with image axes. Advanced deep learning algorithms have significantly improved performance in this field over the past decades. Two-stage and one-stage detectors are two predominant architectures in the realm of object detection, each offering distinct advantages and trade-offs based on their design principles and computational efficiency. On one hands, two-stage detector methods, such as R-CNN (Region-based Convolutional Neural Network) \cite{girshick2015region} and its variants (Fast R-CNN \cite{girshick2015fast}, Faster R-CNN \cite{ren2016faster}), offer higher accuracy by first employing a Region Proposal Network (RPN) and then classifying these proposals into object categories or as background. Two-stage detectors typically follow a coarse-to-fine processing strategy. Initially, the coarse stage focuses on enhancing recall capability, while the refinement stage improves localization based on the initial detection and emphasizes discrimination ability. Although these detectors can achieve high precision without any bells and whistles, they are rarely employed in engineering due to the poor speed and enormous complexity. To speed up the training and inference process, Region-based Fully Convolutional Network (R-FCN) \cite{dai2016r} designs a fully convolutional architecture with shared computation across the entire image, unlike Fast/Faster R-CNN which applies a costly per-region subnetwork multiple times. Conversely, one-stage detectors can retrieve all objects in a single inference step \cite{redmon2016you}. These are popular on mobile devices due to their real-time processing and ease of deployment, but they often struggle with accurately detecting dense and small objects. \cite{liu2016ssd}. Despite its high speed and simplicity, the one-stage detectors have trailed the accuracy of two-stage detectors for years. RetinaNet \cite{ross2017focal} investigates the underlying causes and introduces Focal Loss, modifies the traditional cross-entropy loss to ensure that the detector prioritizes difficult and misclassified examples during training process. This approach makes one-stage detectors achieve  comparable accuracy of two-stage detectors while maintaining a very high detection speed.

\ul{Beyond these paradigms, advancements in traditional object detection for videos and 3D scenes have also contributed valuable insights to the field. TF-Blender} \cite{cui2021tf} \ul{models lower-level temporal relations to increase the feature representation by introducing three modules: temporal relation modelling to preserve spatial information, feature adjustment to enrich neighboring feature maps, and a feature blender to enhance detection performance. This method can be seamlessly integrated into both one-stage and two-stage frameworks to enhance the performance of video object detection. Motion-Aid Feature Calibration Network (MFCN)} \cite{liu2020video} \ul{proposes an end-to-end framework for video object detection  that enhances robustness and efficiency by leveraging optical flow and aggregating features across frames, with R-FCN used as the object detection sub-network.} \cite{cheng2023fusion} \ul{adopts two-stage approaches to design single-modal attacks on camera-LIDAR fusion models for 3D object detection. This method initially identifies vulnerable regions in images under adversarial attacks and then implements tailored attack strategies for various fusion models to generate deployable patches.}

\textbf{Object detection loss function}: In horizontal object detection, loss functions play a crucial role in guiding the training process of deep learning models by measuring the deviation of the predictions and ground truth labels. For classification tasks, the cross-entropy loss and focal loss are widely used. For regression tasks, where precise localization of objects is required, the Smooth L1 loss \cite{girshick2015fast} is commonly employed. This loss function is less sensitive to outliers than the L2 loss, providing stability in training by combining the best properties of L1 and L2 losses. However, Smooth L1 considers the elements of the horizontal bounding box to be independent variables, while eliminates the relationship among them. On the other hand, the Intersection over Union (IoU) loss \cite{he2017mask} is employed to directly optimize the overlap between predicted and ground truth bounding boxes, leading to more precise localization. Following that, various algorithms were introduced to improve IoU loss. G-IoU (Generalized IoU)  \cite{rezatofighi2019generalized}  addressed situations where IoU loss failed to optimize non-overlapping bounding boxes. Distance-IoU (DIoU) \cite{zheng2020distance} improves the IoU by incorporating the normalized distance between the predicted and the ground-truth bounding box. Complete IoU (CIoU) \cite{zheng2020distance} extends DIoU by taking into account three geometric factors: the overlap area, the distance between center points, and the aspect ratio, offering a more comprehensive approach. Although these above methods have been widely used and have achieved adequate performance, horizontal object detectors do not provide accurate orientation when objects appear at arbitrary angles.

\begin{figure*}
    \centering
    \includegraphics[width=1.0\linewidth]{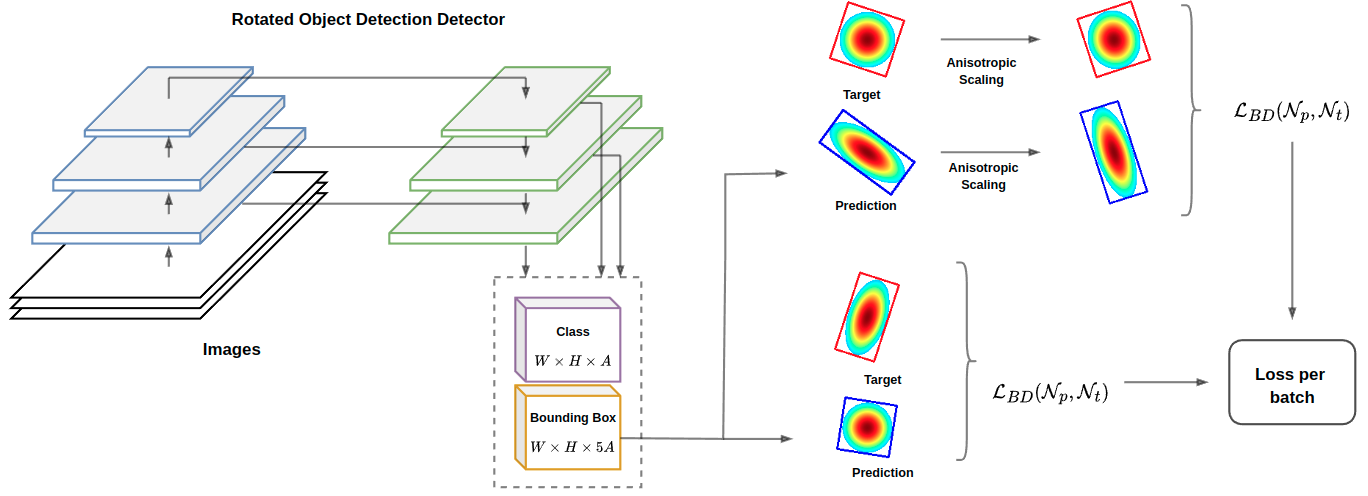}
    \caption{The overall pipeline of our proposed method for rotated object detection problem. 'A' indicates the number of categories.}
    \label{fig:model}
\end{figure*}
\subsection{Rotated Object Detection}
Recent rotated object detectors are highly extended from generic horizontal object detectors with additional angle dimension to represent the oriented object. 

\textbf{Two-stage detector}: numerous outstanding two-stage methods have been proposed for oriented version. The naive Region Proposal Network of RCNN-based model only generates horizontal regions of interest (RoIs), leading to the feature misalignment between horizontal region proposals and rotated bounding boxes. Therefore, the feature representation of object may adversely affected, making the detectors struggle to identify objects and regress precise rotated bounding boxes yet inspiring successive innovations. To address this problem, recent rotated two-stage detector employs rotated region proposal generation and rotated region of interest (RRoI) operators to extract spatial-algined features. For instances, some works proposes Rotated Region Proposal Network (RRPN) \cite{ma2018arbitrary}, which employs rotated anchors to better accommodate objects with various orientations. RRPN generates additional oriented anchors by adding various orientation parameters to horizontal anchors to alleviate the spatial feature misalignment. Therefore, the performance of RRPN is enhanced in terms of recall; however, the redundant rotated anchors bring about expensive computation and memory consumption. To reduce the numbor of rotated anchor boxes, RoI Transformer \cite{ding2019learning} introduces designs lightweight learnable module named RoI Learner, which directly convert horizontal RoIs from naive RPN to rotated RoIs, resulting in better eficiency and accuracy. To make the network architecture simpler without using RoI alignment and regression module, Oriented RPN employs a convolutional block including a 3$\times$3 and two sibling 1$\times$1 convolutional layers to transform HRoIs to RRoIs. Each rotated object is represented using a midpoint offset, which consists of external horizontal bounding boxes and the offset of vertexes with respect to the middle points of the external HBB. Leveraging the lightweight design of Oriented RPN and midpoint offset representation, Oriented RCNN \cite{xie2021oriented} achieves accuracy comparable to state-of-the-art two-stage detectors while also approaching the efficiency levels of one-stage detectors. \ul{ARC} \cite{pu2023adaptive} \ul{improves Oriented R-CNN by incorporating an adaptive rotated convolution module, where convolution kernels dynamically adjust their orientation to align with object orientations in the image. Additionally, an efficient conditional computation method enhances the network's flexibility to capture orientation information for multiple rotated objects, and the module can be seamlessly integrated into any backbone with convolutional layers}.

\textbf{One-stage detector}: Different from two-stage detection frameworks that operate on a coarse-to-fine strategy, one-stage detectors for oriented version perform both classification and regression in a single step. However, one-stage detectors exhibit more severe feature misalignment compared to two-stage due to the removal of RRPN vs RRoI operators. Refined Rotation RetinaNet (R$^3$Det) \cite{yang2021r3det} alleviates this dilemma by initially converting horizontal anchors into rotated anchors. After that, it employs a feature refinement module (FRM) that re-encodes the positional information of the refined bounding box to the relevant feature points using pixel-wise feature interpolation, thereby realizing feature reconstruction and alignment. Similarity, Single-shot Alignment Network (S$^2$ANet) \cite{han2021align} introduces a Feature Alignment Module (FAM) alongside an Oriented Detection Module (ODM). FAM initially generates high-quality anchors using an Anchor Refinement Network module and then adaptively aligns the spatial features according to the corresponding anchor boxes by applying an alignment convolution kernel. Meanwhile, ODM incorporates active rotating filters to encode the orientation information, producing both orientation-sensitive and orientation-invariant features to mitigate the discrepancy between classification scores and localization accuracy. These schemes work in a coarse-to-fine paradigm to align features but are noticeably different from the RRoI operator. The major difference lies in that the FRM or Alignment Convolution follows a full convolution structure and has fewer sampling points than the RRoI operator, making it more efficient. 

\textbf{Anchor-free Rotated Object Detection}: Anchor-free detectors used to eliminate anchor-related hyper-parameters are widely developed, showing potential in the generalization to applications. Existing anchor-free methods for rotated object detection can be divided into two primary categories: keypoint-based approaches and center-based approaches. The keypoint-based methods initially identify a set of adaptive or self-constrained key points, which are then used to outline the spatial boundaries of the object. Oriented Objects Detection Network (O$^2$DNet) \cite{wei2020oriented} first determines the midpoints of four sides of the OBB by regressing the offsets from the center point. Subsequently, it connects two pairs of opposite midpoints to create two mutually perpendicular midlines, which can be decoded to obtain the representation of OBB. Several works extend RepPoints (representative points) \cite{yang2019reppoints} to provide a new finer representation of rotated objects as a set of sample points useful for both localization and recognition. Convex-Hull Feature Adaptation (CFA) \cite{guo2021beyond} proposes convex-hull feature representation to effectively configure convolutional features for oriented and densely packed objects with irregular layouts. Oriented RepPoints \cite{li2022oriented} captures the geometric structure of the objects with sharp variety on orientation in the cluttered environment by employing the adaptive point set of RepPoints as a fine-grained representation instead of directly regressing the angle parameter. To accurately predict the high-quality representation points without requiring points-to-points supervision, Oriented RepPoints designs an Adaptive Points Assessment and Assignment (APAA) modules. This module evaluates the quality of adaptive points, allowing Oriented RepPoints to achieve cutting-edge performances among keypoint-based anchor-free methods.

Center-based methods typically involve generating multiple probabilistic heatmaps to predict a set of candidate points as initial center points, along with a series of feature maps to regress the parameters of oriented bounding boxes \cite{zhao2021polardet, zhang2021arbitrary}. This approach can be largely attributed to the advantages of the anchor-free rotated proposal generation scheme, which not only produces precise proposals but also mitigates the spatial misalignment typically caused by horizontal anchors. However, a notable performance disparity persists between standard center-based oriented methods and other state-of-the-art techniques, underscoring the need for further investigation.

\begin{figure*}[t]
    \centering
    \includegraphics[width=1.0\linewidth]{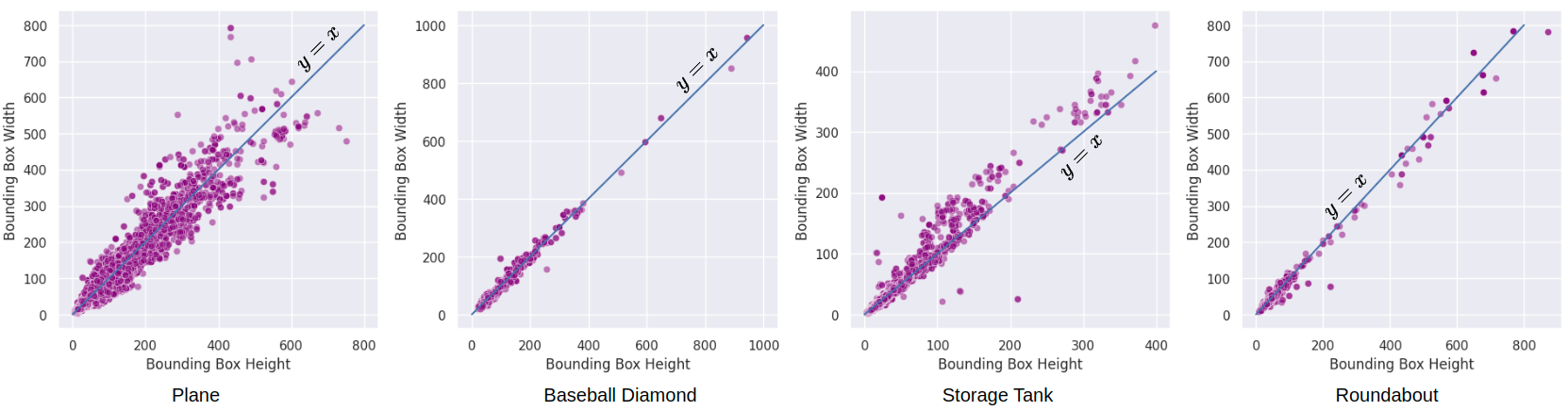}
    \caption{Relationship between bounding box height and width for various object categories on DOTA-v1.0 dataset. The clustering of data points along the $y = x$ line indicates a significant presence of square-like objects in the dataset.}
    \label{fig:scatter}
\end{figure*}

\textbf{Regression Loss for Rotated Object Detection}: Several works extend the $l_n$ loss function used in traditional object detection for rotated case. However, the $l_n$-based loss often encounters issues such as boundary discontinuity and square-like problem, attributed to the periodic nature of angle parameters and the variability in bounding box definitions.  Additionally, there exists an inconsistency between the metric and $l_n$ loss, wherein a lower training loss does not necessarily translate to improved performance. Although IoU loss and its variants (e.g. GIoU \cite{rezatofighi2019generalized}, DIoU \cite{zheng2020distance}, CIoU \cite{zheng2020distance}) can align the object detection metric with the loss, they  are not directly applicable to rotated detectors due to the indifferentiable nature of rotating IoU. Therefore, several methods proposes differentiable functions to approximate IoU loss between two rotated bounding boxes. For example, PIoU \cite{chen2020piou} simply counts the number of overlapping pixels using a differentiable kernel function. Other works try to integrated rotating IoU as a loss weight of regression loss \cite{yang2019scrdet, yang2022scrdet++}. The $l_n$-norm loss is used to control the direction of gradient propagation, while rotating RIoU parameter adjust gradient magnitude.

Recent works introduce a cohesive and sophisticated solution to address the issues of boundary discontinuity and the square-like problem by utilizing Gaussian distribution. The classical oriented bounding box representation $\mathcal{B}(x, y, w, h, \theta)$ is transformed to a bivariate Gaussian distribution $\mathcal{N}(\mu, \Sigma)$ where mean $\mu=(x, y)$ denotes the object center, and the covariance matrix $\Sigma^{1/2}=RSR^T$ where R and S are the rotation matrix and diagonal matrix of eigenvalues, respectively. The computations of R and S are defined as following:
\begin{align}
    R=\begin{bmatrix}
        cos\theta & -sin\theta \\
        sin\theta & cos\theta
    \end{bmatrix}, \ S = \begin{bmatrix}
        \frac{h}{2} & 0 \\
        0 & \frac{w}{2}
    \end{bmatrix}
\end{align}

The  advantage of using Gaussian distribution is that the angle is encoded by trigonometric function thereby not constrained by periodicity of angle. Moreover, the OBB parameters are joint-optimized dynamically so that they can influence each other during training. Several distance measures are employed to compare two multivariate Gaussian distribution, including Generalize Wastersein Distance (GWD)\cite{yang2021rethinking}, Kullback-Leiber Divergence (KLD) \cite{yang2021learning}, and Kalman Filter-based IoU \cite{yang2022kfiou}. 
\section{Methodology}
\label{sec:proposed_method}

\subsection{Gaussian Representation for Bounding Box}
According to \cite{yang2021rethinking}, to prevent the boundary discontinuity and square-like problems, the arbitrary-oriented bounding box $\mathcal{B}(x, y, w, h, \theta)$ is converted into bivariate Gaussian distribution $\mathcal{N}(\bm{\mu}, \mathbf{\Sigma})$ using the following formula:
\begin{align*}
    \bm{\mu} &= [x, y]^T \\
    \mathbf{\Sigma}^{\frac{1}{2}} 
    &= \mathbf{RSR}^T \\
    &= \begin{bmatrix}
        cos\theta & -sin\theta \\
        sin\theta & cos\theta
    \end{bmatrix} \begin{bmatrix}
        \frac{w}{2} & 0 \\ 
        0 & \frac{h}{2}
    \end{bmatrix} \begin{bmatrix}
        cos\theta & sin\theta \\ 
        -sin\theta & cos\theta
    \end{bmatrix} \\
    &= \begin{bmatrix}
        \frac{w}{2}cos^2\theta + \frac{h}{2}sin^2\theta & \frac{w-h}{2}cos\theta sin\theta \\
        \frac{w-h}{2}cos\theta sin\theta & \frac{w}{2}sin^2\theta + \frac{h}{2}sin^2\theta
    \end{bmatrix}
\end{align*}
where $\mathbf{R}$ and $\mathbf{S}$ denote the rotation matrix and square diagonal matrix, respectively. The covariance matrix $\mathbf{\Sigma}$ is computed as follow:
\begin{align*}
    \mathbf{\Sigma} &= \mathbf{RS}^2\mathbf{R}^T \\&= \begin{bmatrix}
        \frac{w^2}{4}cos^2\theta + \frac{h^2}{4}sin^2\theta & \frac{w^2-h^2}{4}cos\theta sin\theta \\
        \frac{w^2-h^2}{4}cos\theta sin\theta & \frac{w^2}{4}sin^2\theta + \frac{h^2}{4}cos^2\theta
    \end{bmatrix}
\end{align*}

After that, Gaussian distances have been utilized to measure the deviation between two oriented bounding boxes, such as GWD\cite{yang2021rethinking}, KLD\cite{yang2021learning}. The bivariate Gaussian representation of bounding boxes has several properties to address some problems for rotated object detection loss computation:

\begin{itemize}
    \item Property 1. $\mathbf{\Sigma}(w, h, \theta)=\mathbf{\Sigma}(h, w, \theta-\frac{\pi}{2})$: This property ensures that both the OpenCV and long-edge definitions are equivalent when using Gaussian-based distances.
    \item Property 2. $\mathbf{\Sigma}(w, h, \theta)=\mathbf{\Sigma}(w, h, \theta-\pi)$: two bounding boxes $\mathcal{B}(x,y,w,h,\theta)$ and $\mathcal{B}(x,y,w,h,\theta-\pi)$ have the similar Gaussian representation, eliminating the boundary discontinuity problem.
    \item Property 3. $\mathbf{\Sigma}(w, h, \theta)\approx \mathbf{\Sigma}(w, h, \theta-\pi/2)$, if $w\approx h$: for square-like bounding boxes, the Gaussian representations are the same when box is rotated by $\frac{\pi}{2}$, $\pi$, and $\frac{3\pi}{2}$, preventing the square-like problem.
\end{itemize}

\begin{figure}[htbp]
    \centering
    \includegraphics[width=0.95\linewidth]{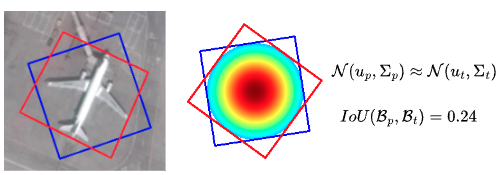}
    \caption{Example of isotropic Gaussian case. Both square-like \textcolor{red}{red} (ground-truth) and \textcolor{blue}{blue} (prediction) bounding boxes represent the same Gaussian distribution.}
    \label{fig:isotropic}
\end{figure}

However, for square-like object, the variance along each dimension is equal, therefore the bounding box represents the same Gaussian distribution in all directions. For example, for two square-like bounding boxes $\mathcal{B}(x, y, w, h, \theta)$ and $\mathcal{B}(x, y, w, h, \theta + \frac{\pi}{4})$ with $w\approx h$, their Gaussian representations are similar, thus the Gaussian-based distances are approximate to 0. However, the Intersection over Union distance between two bounding boxes are noticeable (see Figure \ref{fig:isotropic}). To eliminate the isotropic Gaussian problem, the covariance matrix of Gaussian distribution representation of square-like bounding box is adjusted to:

\begin{align*}
    \mathbf{\Sigma}^{1/2}=\begin{bmatrix}
        cos4\theta & -sin4\theta \\
        sin4\theta & cos4\theta
    \end{bmatrix}\begin{bmatrix}
        \frac{h'}{2} & 0 \\ 
        0 & \frac{w'}{2}
    \end{bmatrix}\begin{bmatrix}
        cos4\theta & sin4\theta \\
        -sin4\theta & cos4\theta
    \end{bmatrix}
\end{align*}

where $h'=h(1+\frac{cos4\theta}{\delta})$ and $w'=w(1-\frac{cos4\theta}{\delta})$ are new eigenvalues. For two square-like boxes, the new representation satisfies all above properties:

\begin{itemize}
    \item For square-like case, $w\approx h$, $cos(4\theta)=cos(4(\theta-\frac{\pi}{2}))$, the rotation matrices $R$ and eigenvalue matrices $S$ of two bounding-boxes are similar, therefore $\mathbf{\Sigma'}(w, h, \theta)\approx \mathbf{\Sigma'}(w, h, \theta-\frac{\pi}{2})\approx \mathbf{\Sigma'}(h, w, \theta -\frac{\pi}{2})$, so the Property 1 and Property 3 are satisfied.
    \item Since $cos(4\theta)=cos(4(\theta-\pi))$, therefore $\mathbf{\Sigma'}(w, h, \theta)\approx \mathbf{\Sigma'}(w, h, \theta-\pi)$, satisfying Property 2.
\end{itemize}
In addition, the new representation ensures that $\mathbf{\Sigma'}(w, h, \theta)\neq \mathbf{\Sigma'}(w, h, \theta + \theta')$ where $\theta' \notin \{k\frac{\pi}{2} | k\in Z\}$. The use of anisotropically scaling during training process is illustrated in Figure \ref{fig:model}. In our experiments, we set $\sigma=5$ to ensure that the proposed loss aligns with the IoU-based loss for square-shaped bounding boxes.

The scatter plot in Figure \ref{fig:scatter} illustrates significant presence of square-shaped bounding boxes of four different object categories of DOTA dataset: Plane, Baseball Diamond, Storage Tank, and Roundabout (note that other categories in the dataset also contain square-like objects, but to a lesser extent). For the given object categories, the bounding boxes typically possess similar height and width dimensions, confirming their square-like properties.

\subsection{Distance between two bounding boxes}
Although Generalized Wasserstein Distance (GWD) \cite{piccoli2014generalized} and Kullback-Leiber Divergence (KLD) \cite{kullback1951information} can measure the deviation between two multivariate Gaussian distribution, these have drawbacks for object detection problem. \cite{yang2021learning} shows several disadvantages of GWD, specially focus on scale variance nature of GWD. For Kullback-Leiber Divergence, it has some differences compared to IoU-based metrics. The KLD between two bivariate Gaussian distribution is defined as:

\begin{align*}
    D_{KL}(\mathcal{N}_p\|\mathcal{N}_t) &= \frac{1}{2}(\bm{\mu}_p-\bm{\mu}_t)^T
    \mathbf{\Sigma}_t^{-1}(\bm{\mu}_p-\bm{\mu}_t) \\
    &+ \frac{1}{2}tr(\mathbf{\Sigma}_t^{-1}\mathbf{\Sigma}_p) - \frac{1}{2}ln\frac{|\mathbf{\Sigma}_p|}{|\mathbf{\Sigma}_t|} - 1
\end{align*}

The major disadvantage of KLD is its asymetric nature, meaning $D_{KL}(P||Q) \neq D_{KL}(Q||P)$. Asymmetric loss functions inherently introduce a bias towards certain types of errors. For example, they may penalize overestimation more heavily than underestimation or vice versa. This bias can lead to suboptimal performance, particularly if the nature of errors or their impact is not uniform across different object detection scenarios. In contrast, the Bhattacharyya distance is symmetric, thus it is more natural in the context of IoU. Furthermore, the Bhattacharyya distance is designed to measure the amount of overlap between two probability distributions. It's particularly effective at recognizing and quantifying partial overlaps, which aligns well with the concept of IoU that measures the overlap between predicted and ground-truth regions.

The Bhattcharyya distance between two bivariate Gaussian distribution $\mathcal{N}(\bm{\mu}_p, \mathbf{\Sigma}_p)$ and $\mathcal{N}(\bm{\mu}_t, \mathbf{\Sigma}_t)$:
\begin{align*}
        D_B &= \alpha \frac{1}{8} (\bm{\mu}_p - \bm{\mu}_t)^T\mathbf{\Sigma}^{-1}(\bm{\mu}_p -\bm{\mu}_t)  \\ 
        &+ \frac{1}{2} ln\frac{det(\mathbf{\Sigma})}{det(\mathbf{\Sigma_p}^{1/2})det(\mathbf{\Sigma}^{1/2})}
\end{align*}
where  $\mathbf{\Sigma}=(\mathbf{\Sigma}_p + \mathbf{\Sigma}_t)/2$ is the average of two covariance matrices. The first term (mean different term) is squared Mahalanobis distance\cite{mahalanobis2018generalized}, captures the distance between two points in a multivariate space, taking into account the correlations between variables. The second term (covariance similarity term) measures the similar of covariance matrix which representing the shape and size of rotated bounding boxs. In the object detection context, the covariance matrices often are large (i.e., the distributions are very spread out in the feature space), hence the inverse of average covariance matrix will have small values, leading the Mahalanobis term will be relatively small. Therefore, we increase the first term by $\alpha$ coefficient. To ensure the proposed loss function aligns with IoU loss, we compared it with IoU loss and determined that setting $\alpha = 3$ achieves the desired alignment. The final proposed loss is defined as:
\begin{equation}
    \mathcal{L}_{BD}(\mathcal{B}_p, \mathcal{B}_t) = 1 - \frac{1}{1 + \sqrt{D_B(\mathcal{N}_p, \mathcal{N}_t))}}
\end{equation}
\subsection{Consistent with IoU-based distance}
While the Intersection over Union (IoU) and Bhattacharyya distance are both measures used to evaluate similarity or overlap, finding the direct mathematical relationship between them is non-trivial task. Alternatively, we demonstrate that the Bhattacharyya distance satisfies all the desirable properties of an IoU-based distance metric. Two appealing features that make IoU distance widely used for evaluating various 2D/3D computer vision tasks are as follows:
\begin{itemize}
    \item IoU as a loss function is a metric. This means that IoU loss ($\mathcal{L}_{IoU} = 1 - IoU$) satisfies all the properties of a metric, including non-negativity, identity of indiscernibles, symmetry, and the triangle inequality.
    \item IoU is scale-invariant, meaning the similarity between two arbitrary shapes, A and B, remains unaffected by the scale of their space
\end{itemize}

\begin{figure*}[htbp]
    \centering
    \includegraphics[width=1.0\linewidth]{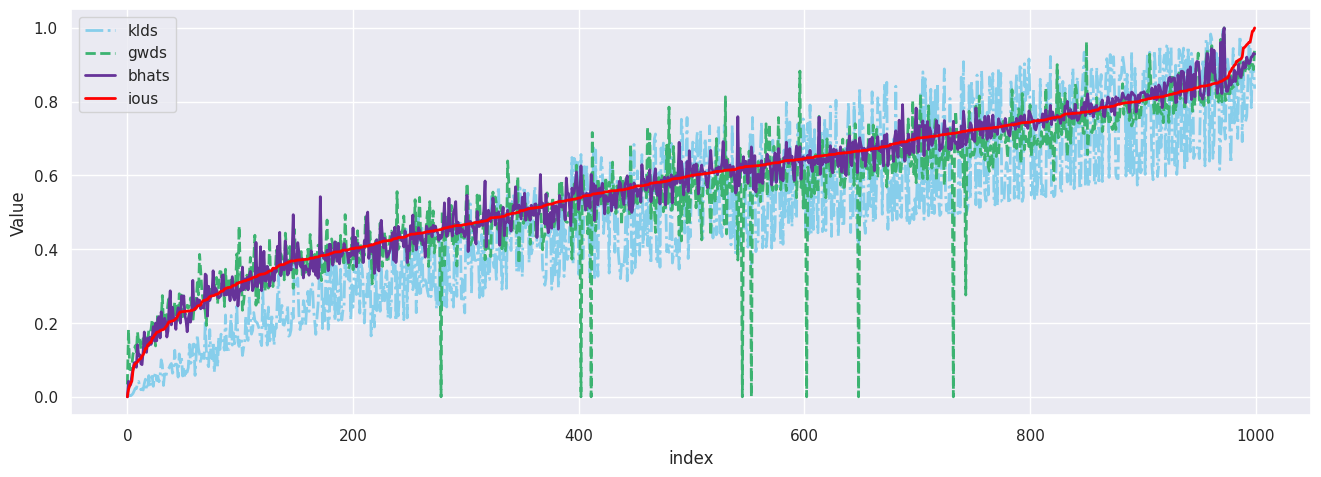}
    \caption{Comparison of different loss functions over 1000 randomized horizontal bounding boxes pairs. Notably, the Bhattacharyya Loss closely follows the trend of the Complete IoU Loss, indicating their equivalence and similar performance characteristics, whereas GWD Loss and KLD Loss exhibit higher variability and diverge more from the Complete IoU Loss.}
    \label{fig:loss_comparison}
\end{figure*}
In this section, we provide a proof to show $\mathcal{L}_{BD}$ between two Gaussian Bounding Boxes is a distance and holds all properties of a metric, including non-negativity, identity of indiscernibles, symmetry and triangle inequality.

\subsubsection{Non-negativity}
\textbf{Proposion 1}: For any two Gaussian Bounding Boxes, the Bhattacharyya distance loss function between them is non-negative, $i.e\  \forall \mathcal{N}_1(\mu_1, \Sigma_1),\ \mathcal{N}_1(\mu_1, \Sigma_1),\ \mathcal{L}_{DB}(\mathcal{N}_1, \mathcal{N}_2) \geq 0$.

\textbf{Proof 1}: Because Bhattacharyya distance is always non-negative, therefore $\frac{1}{1 + \sqrt{D_B}} \leq 1$. Thus, $\mathcal{L}_{DB} \geq 0$.

\subsubsection{Identity of indiscernibles}
\textbf{Proposion 2}: The Bhattacharyya distance loss function between two Gaussian Bounding Boxes is zero if and only if they are identical, $i.e\ \mathcal{L}_{BD}(\mathcal{N}_1, \mathcal{N}_2)=0 \Leftrightarrow \mathcal{N}_1 = \mathcal{N}_2$.

\textbf{Proof 2}: If $\mathcal{N}_1=\mathcal{N}_2$, $\mu_1 = \mu_2$, $\Sigma_1 = \Sigma_2$, $\Sigma = (\Sigma_1 + \Sigma_2)/2 = \Sigma_2$, therefore both term of Bhattacharyya distance is equal to 0, thus Bhattacharyya distance loss is 0. Consequently, $\mathcal{N}_1=\mathcal{N}_2\Rightarrow \mathcal{L}_{BD}(\mathcal{N}_1, \mathcal{N}_2)=0$.

if $\mathcal{L}_{BD}(\mathcal{N}_1, \mathcal{N}_2)=0$, both mean different and covariance similarity terms are equal to zero. 
For mean different term $(\mu_1 - \mu_2)^T \Sigma^{-1} (\mu_1 - \mu_2) \Rightarrow \mu_1 = \mu_2$ because $\Sigma^{-1}$ is positive definite matrix.
For all $\lambda \in [0, 1]$, the multiplicative form of Brunn–Minkowski inequality states that:
\begin{align*}
    &det(\lambda \Sigma_1 + (1-\lambda)\Sigma_2) \geq det(\Sigma_1)^\lambda det(\Sigma_2)^{1-\lambda} \\
    \Rightarrow\  
    &det(\frac{\Sigma_1 + \Sigma_2}{2}) \geq det(\Sigma_1)^{1/2}det(\Sigma_2)^{1/2}
\end{align*}

The equality holds in the Brunn-Minkowski inequality if and only if $\Sigma_1 = k\Sigma_2$ ($k > 0$ due to both $\Sigma_1$ and $\Sigma_2$ are positive definite matrices). Equality holds when:
\begin{align*}
    &det(\lambda \Sigma_1 + (1-\lambda)\Sigma_2) = det(\Sigma_1)^\lambda det(\Sigma_2)^{1-\lambda} \\
    \Leftrightarrow\  
    &det(\frac{1 + k}{2} \Sigma_2) = det(k\Sigma_2)^{1/2} det(\Sigma_2)^{1/2} \\
    \Leftrightarrow\  
    &\left (\frac{1+k}{2} \right )^n det(\Sigma_2) = \sqrt{k^n}det(\Sigma_2) \\
    \Leftrightarrow\  
    &\left (\frac{1+k}{2} \right )^n = \sqrt{k^n} \\
    \Leftrightarrow\
    &k = 1\  (when\ n = 2) \Leftrightarrow\  \Sigma_1 = \Sigma_2
\end{align*}
where $n$ denotes the dimensionality of the space. In the context of rotated object detection, $n=2$, therefore $\Sigma_1 = \Sigma_2$. Consequently, $\mathcal{L}_{BD}(\mathcal{N}_1, \mathcal{N}_2)=0 \Rightarrow \mathcal{N}_1=\mathcal{N}_2$.
\subsubsection{Symmetry}
\textbf{Proposion 3}: Bhattachayya Distance loss is a symmetric function, $i.e\  \mathcal{L}_{BD}(\mathcal{N}_1, \mathcal{N}_2) = \mathcal{L}_{BD}(\mathcal{N}_2, \mathcal{N}_1)$ for any two Gaussian Bounding Boxes $\mathcal{N}_1(\mu_1, \Sigma_1)$ and $\mathcal{N}_2(\mu_2, \Sigma_2)$.

\textbf{Proof 3}: Since the Bhattacharyya distance between two multivariate Gaussian distributions is symmetric, $\mathcal{L}_{BD}$ inherits this symmetry.

\subsubsection{Triangle inequality}
\textbf{Proposion 4}: For any three Gaussian Representations of rotated bounding boxes $\mathcal{N}_1(\mu_1, \Sigma_1),\ \mathcal{N}_2(\mu_2, \Sigma_2)$, and $\mathcal{N}_3(\mu_3, \Sigma_3)$, triangle inquality holds true:
\begin{align*}
    \mathcal{L}_{BD}(\mathcal{N}_1, \mathcal{N}_3) \leq
    \mathcal{L}_{BD}(\mathcal{N}_1, \mathcal{N}_2) +
    \mathcal{L}_{BD}(\mathcal{N}_2, \mathcal{N}_3)
\end{align*}

\begin{figure}[htbp]
    \centering
    \includegraphics[width=1.0\linewidth]{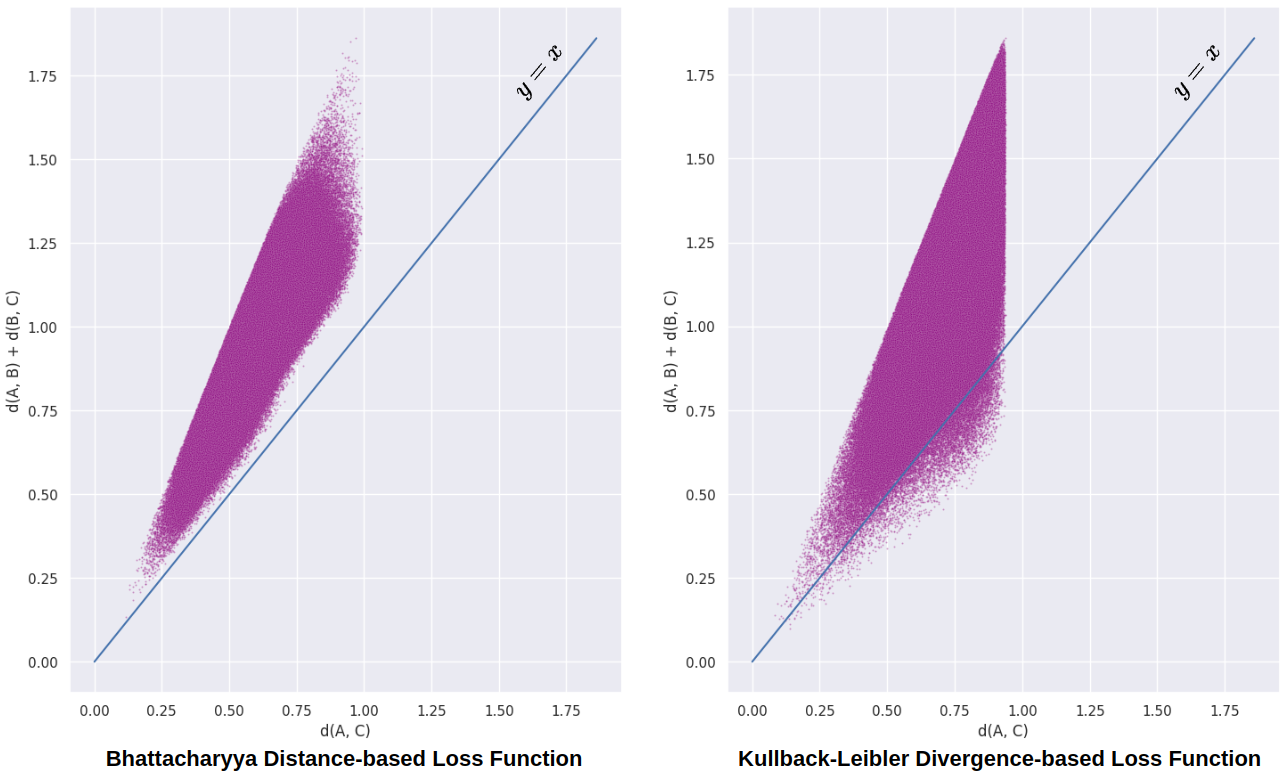}
    \caption{Scatter Plot of Triangle Inequality Property of $\mathcal{L}_{BD}$ and $\mathcal{L}_{KLD}$. $d(\cdot, \cdot)$ denotes loss function. Points under the line $y=x$ indicates that the corresponding loss does not satisfy triangle inequality.}
    \label{fig:triangle}
\end{figure}
\textbf{Proof 4}: Following \cite{rezatofighi2019generalized}, the correctness of the proposition is checked by evaluating several random samples. In this experiment, we sample three rotated bounding boxes over $10^6$ iterations and convert them to Gaussian Representations, denoted at $\mathcal{N}_1, \mathcal{N}_2$, and $\mathcal{N}_3$. For each iterations, we compute the Bhattacharyya Distance loss for each pair of elements in the randomly chosen set of three bounding boxes and find the maximum loss value, e.g. $\mathcal{L}_{BD}(\mathcal{N}_1, \mathcal{N}_3) \geq \mathcal{L}_{BD}(\mathcal{N}_1, \mathcal{N}_2)$ and $\mathcal{L}_{BD}(\mathcal{N}_1, \mathcal{N}_3) \geq \mathcal{L}_{BD}(\mathcal{N}_2, \mathcal{N}_3)$. By checking whether the sum of the two smaller losses exceeds or equals the largest loss, we assess the adherence of the loss function to the triangle inequality condition. Throughout all iterations, the condition $\mathcal{L}_{BD}(\mathcal{N}_1, \mathcal{N}_3) \leq
    \mathcal{L}_{BD}(\mathcal{N}_1, \mathcal{N}_2) +
    \mathcal{L}_{BD}(\mathcal{N}_2, \mathcal{N}_3)$ held. By applying above procedure to Generalize Wasserstein Distance and Kullback-Leibler Divergence loss, we observe that $\mathcal{L}_{GWD}$ satisfies the triangle inequality, but $\mathcal{L}_{KLD}$ does not fulfill this property (as illustrated by scatter plot in Figure \ref{fig:triangle}.
    
\subsubsection{Scale-invariant}
\textbf{Proposion 5}: The Bhattacharyya Distance loss function is invariant to the scale of the problem.

\textbf{Proof 5}: Let transform two rotated bounding boxes $\mathcal{B}_p$ and $\mathcal{B}_t$ using a transformation matrix $M \in \mathbb{R}^{2\times 2}$, the converted Gaussian representation are $\mathcal{N}'_p(M\mu_p, M\Sigma_pM^T)$ and $\mathcal{N}'_t(M\mu_t, M\Sigma_tM^T)$. The new mean covariance matrix is:
\begin{align*}
\Sigma'&=\frac{M\Sigma_pM^T+M\Sigma_tM^T}{2} \\&= M\frac{\Sigma_p + \Sigma_t}{2}M^T = M\Sigma M^T
\end{align*}

The Bhattacharyya distance between two transformed bounding boxes is:
\begin{align*}
    &D_B(\mathcal{N}'_p, \mathcal{N}'_t) \\
    &=\frac{1}{8}(\mu_p-\mu_t)^TM^T(M^T)^{-1}\Sigma^{-1}M^{-1}M(\bm{\mu}_p-\bm{\mu}_t) \\
    &+ \frac{1}{2}ln\frac{|M\Sigma M^T|}{|M\Sigma_p M^T|^{1/2}|M\Sigma_t M^T|^{1/2}} = D_B(\mathcal{N}_p, \mathcal{N}_t)
\end{align*}

Therefore the Bhattacharyya distance-based loss ensures the scale-invariant property. \cite{yang2021learning} shows that KLD loss also obeys the scale-invariant property, while GWD loss does not. In summary, $\mathcal{L}_{BD}$ holds all the major properties of IoU-based loss function, while $\mathcal{L}_{GWD}$ does not satisfy scale-invariant, and $\mathcal{L}_{KLD}$ does not obey triangle-inequality and symmetry properties (as shown in Table \ref{tab:properties}).

\begin{table}[htbp]
    \centering
    \renewcommand{\arraystretch}{1.5}
    \caption{A comparison of different loss functions regarding their properties. Our proposed loss possesses all the appealing properties of the IoU-based loss function.}
     \resizebox{\columnwidth}{!}{%
     \begin{tabular}{|p{1.5cm}|>{\centering}p{1.2cm}|>{\centering}p{1.8cm}|>{\centering}p{1.3cm}|>{\centering}p{1.2cm}|>{\centering\arraybackslash}p{1.2cm}|}
        \hline
         Loss & Non-negativity & Identity of indiscernibles & Symmetry & Triangle-inequality & Scale-invariant\\ \hline
         $\mathcal{L}_{GWD}$ \cite{yang2021rethinking} & \checkmark & \cmark & \cmark & \cmark & \xmark \\
         $\mathcal{L}_{KLD}$ \cite{yang2021learning}& \cmark & \cmark & \xmark & \xmark & \cmark \\ 
         $\mathcal{L}_{IoU}$ \cite{he2017mask} & \cmark & \cmark & \cmark & \cmark & \cmark \\
         $\mathcal{L}_{BD}$ (ours) & \cmark & \cmark & \cmark & \cmark & \cmark \\
          \hline
    \end{tabular}
    }
    \label{tab:properties}
\end{table}

Figure \ref{fig:loss_comparison}  illustrates a detailed comparison between three loss functions (GWD, KLD, and Bhattacharyya Distance Loss) and the state-of-the-art horizontal regression loss Complete IoU (CIoU) over 1000 pairs of randomized horizontal bounding boxes. Throughout the 1000 iterations, GWD and KLD Loss (depicted in lightblue and lightgreen color, respectively) exhibit a higher degree of variability. Their values oscillate significantly, suggesting that these loss functions may not provide as consistent feedback for model training in the object detection context. In contrast, the Bhattacharyya Loss and the Complete IoU Loss (shown in purple and red color, respectively) appear much smoother and more stable over the iterations. This stability is indicative of a more reliable performance in guiding the model training process. 

\begin{figure}[htbp]
    \centering
    \includegraphics[width=1.0\linewidth]{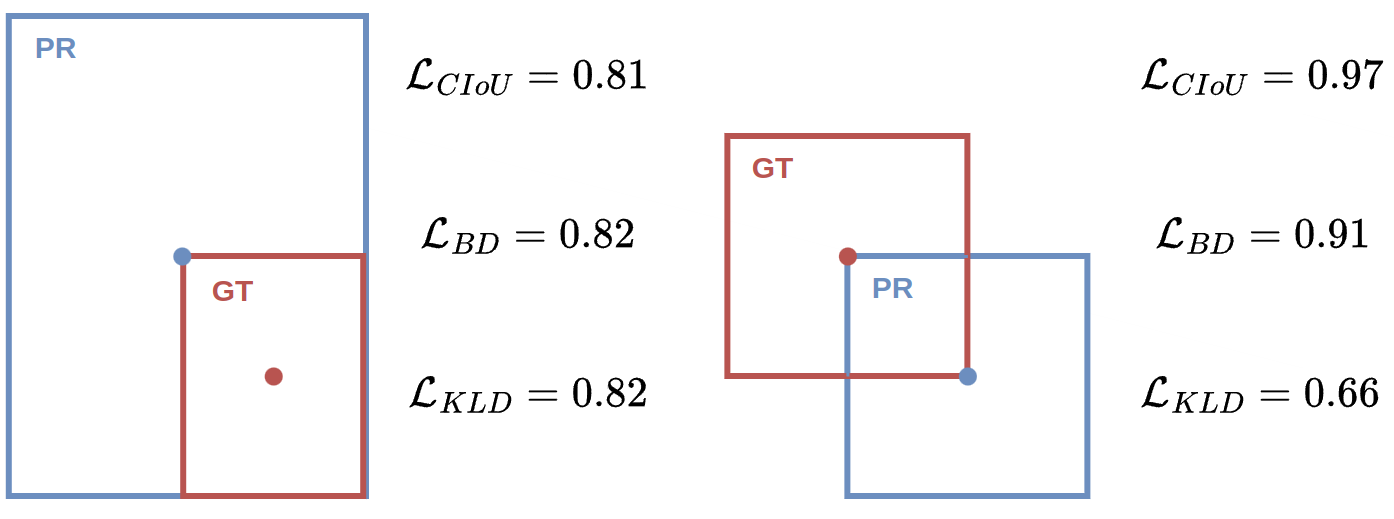}
    \caption{Comparison of loss metrics for horizontal bounding box overlap. \textbf{\textcolor{red}{Red}} and \textbf{\textcolor{blue}{blue}} indicate ground-truth (GT) and predicted (PR) bounding boxes, respectively.}
    \label{fig:iou_kl_bhat}
\end{figure}

A key observation from the plot is that the Bhattacharyya Loss closely mirrors the trend of the Complete IoU Loss. The two lines almost overlap for the majority of the iterations, which suggests that the Bhattacharyya Loss is effectively equivalent to the Complete IoU Loss in terms of performance characteristics. This equivalence implies that either loss function could be employed with similar expected outcomes in model training scenarios. On the other hand, the GWD Loss and KLD Loss diverge more significantly from the Complete IoU Loss. The higher variability and deviations highlight these losses as potentially less optimal for this specific task when compared to the Bhattacharyya Loss and Complete IoU Loss. Thus, the plot underscores the relative consistency and reliability of the Bhattacharyya and Complete IoU Losses over the less stable GWD and KLD Losses. Additionally, the increase in Bhattacharyya Distance loss suggests that it maintains more consistency with the increasing overlap indicated by CIoU. Meanwhile, Kullback-Leiber Divergence decreases, showing a differing trend relative to the increases in both CIoU and Bhattacharyya Distance Losses (shown in Figure \ref{fig:iou_kl_bhat}).

\subsection{Overall Object Detection Framework}
To incorporate the proposed regression loss, we employ two common rotated object detectors, RetinaNet \cite{ross2017focal} and R3Det \cite{yang2021r3det}. Both detectors are one-stage object detection architecture known for achieving a good balance between speed and accuracy. 

\textbf{Encoder}: These detectors typically uses pre-trained models like ResNet (ResNet-50 or ResNet-101) as its encoder or backbone. The encoder extracts hierarchical feature maps from the input image at multiple scales. For instance, $F_i,\ i \in \{1, 2, 3, 4, 5\}$ are feature extracted from ResNet. The resolution of $F_i$ is $\frac{H}{2^i} \times \frac{W}{2^i} \times C_i$ where $C_i$ denotes the number of channels at level $i$.

\textbf{Feature Pyramid Network (FPN)}: The Feature Pyramid Network \cite{lin2017feature} combines high-resolution (low-level features) with low-resolution (high-level features) to create a feature pyramid. Outputs from the backbone $F_3, F_4, F_5$ are processed to create feature maps $P_3, P_4, P_5$ and additional levels $P_6, P_7$, where $P_6$ is generated by applying a stride-2 convolution to $F_5$, and $P_7$ is derived from $P_6$ using another stride-2 convolution. These feature maps ($P_3$ to $P_7$) capture semantic information at different scales. Similar to the outputs of encoder, the resolution of $P_i$ is $\frac{H}{2^i} \times \frac{W}{2^i} \times C_i$.

\textbf{Classification and Regression Head}: RetinaNet and R3Det generate rotated anchors for each spatial location on the feature maps and use two separate heads: classification head to predicts class probabilities for each anchor ($C_i=\{c_1, c_2,..., c_{Ncls+1}\} \in \mathbb{R}^{Ncls+1}$ where $N_{cls}$ is number of categories and $c_i$ is class probabilities) and regression head to predicts bounding box for each anchor ($A_i=(x_i, y_i, w_i, h_i, \theta_i) \in \mathbb{R}^5$). These heads are lightweight, sharing the same architecture across all feature levels.

\textbf{Loss Function}: The loss function in RetinaNet is central to its performance, especially its ability to handle the class imbalance between background and foreground objects. In our experiment, we combine focal loss for classification and Bhattacharyya Distance-based loss function for bounding box regression. The total loss is a linear combination of classification loss and regression loss for all anchors:
\begin{align*}
    \mathcal{L}_{total} = \frac{1}{N_{pos}} \left( \sum_{i=1}^N \mathcal{L}_{focal}(C_i) + \lambda \sum_{j=1}^{N_{pos}} \mathcal{L}_{BD}(A_i, G_i)\right) 
\end{align*}  
where $N_{pos}$ is number of positive anchors by computing during the training process, each anchor is defined as positive if the IoU between it and any ground-truth box is greater than a defined threshold (e.g. 0.5). $\mathcal{L}_{focal}(C_i)$ evaluates the discrepancy between the predicted class probabilities and the true class labels.  $G_i$ is matched ground-truth for anchor bounding box $A_i$. Both $G_i$ and $A_i$ are converted to multivariate Gaussian distances before computing regression loss; if $G_i$ is squared-like shape, both are anisotropic scaled. $\lambda$ is a scaling factor to balance the two loss terms, which is set to $2.0$ in our experiments.
\begin{table*}[t]
    \renewcommand{\arraystretch}{1.5}
    \caption{Evaluation on DOTA-1.0 test set. The evaluation metric is mean AP\(_{50}\) and AP\(_{50}\) per category.}
    \centering
    \scalebox{0.77}{
    \begin{tabular}{|c|c|c|c|c|c|c|c|c|c|c|c|c|c|c|c|c|c|c|}
        \hline
         \textbf{Model} & \textbf{Loss} & \textbf{PL} & \textbf{BD} & \textbf{BR} & \textbf{GFT} & \textbf{SV} & \textbf{LV} & \textbf{SH} & \textbf{TC} & \textbf{BC} & \textbf{ST} & \textbf{SBF} & \textbf{RA} & \textbf{HA} & \textbf{SP} & \textbf{HC} & $\textbf{AP}_{50}$ \\
         \hline
         \multirow{5}{*}{\shortstack{RetinaNet\\ \cite{ross2017focal}}} 
         & $\mathcal{L}_{SmoothL1}$ & \textcolor{brightgreen}{\textbf{\underline{89.41}}} & 76.83 & 40.90 & 67.57 & 77.51 & 62.67 & 77.54 & \textcolor{red}{\textbf{90.89}} & 82.34 & 81.99 & \textcolor{brightgreen}{\textbf{\underline{58.16}}} & \textcolor{red}{\textbf{61.56}} & 56.46 & 63.70 & 38.96 & 68.43 \\ 
         & $\mathcal{L}_{GWD}$ & 88.57 & 77.88 & 41.35 & 71.06 & 78.22 & 68.37 & 84.13 & \textcolor{brightgreen}{\textbf{\underline{90.90}}} & \textcolor{brightgreen}{\textbf{\underline{84.71}}} & \textcolor{red}{\textbf{82.24}} & 55.41 & \textcolor{brightgreen}{\textbf{\underline{63.87}}} & \textcolor{red}{\textbf{59.49}} & 63.86 & 40.99 & 70.07 \\ 
         & $\mathcal{L}_{KLD}$ & 89.01 & \textcolor{brightgreen}{\textbf{\underline{79.68}}} & \textcolor{red}{\textbf{42.66}} & 72.40 & \textcolor{red}{\textbf{78.72}} & \textcolor{red}{\textbf{69.14}} & 84.46 & 90.84 & 83.51 & 80.48 & 53.89 & 61.47 & 57.97 & \textcolor{brightgreen}{\textbf{\underline{68.63}}} & 41.80 & \textcolor{red}{\textbf{70.31}} \\ 
         & $\mathcal{L}_{KFIoU}$ & \textcolor{red}{\textbf{89.35}} & 76.38 & 41.98 & \textcolor{brightgreen}{\textbf{\underline{74.29}}} & 78.18 & 68.44 & \textcolor{red}{\textbf{84.51}} & \textcolor{red}{\textbf{90.89}} & 83.31 & 82.02 & 52.62 & 60.05 & 59.12 & 64.89 & \textcolor{red}{\textbf{43.35}} & 69.96 \\ 
          & $\mathcal{L}_{BD}$ & 88.97 & \textcolor{red}{\textbf{78.69}} & \textcolor{brightgreen}{\textbf{\underline{43.18}}} & \textcolor{red}{\textbf{{73.05}}} & \textcolor{brightgreen}{\textbf{\underline{78.75}}} & \textcolor{brightgreen}{\textbf{\underline{72.05}}} & \textcolor{brightgreen}{\textbf{\underline{85.98}}} & 90.85 & \textcolor{red}{\textbf{84.43}} & \textcolor{brightgreen}{\textbf{\underline{82.98}}} & \textcolor{red}{\textbf{57.62}} & 61.30 & \textcolor{brightgreen}{\textbf{\underline{62.52}}} & \textcolor{red}{\textbf{68.33}} & \textcolor{brightgreen}{\textbf{\underline{49.26}}} & \textcolor{brightgreen}{\textbf{\underline{71.86}}} \\ 
         \hline

         \multirow{5}{*}{\shortstack{R3Det\\ \cite{yang2021r3det}}} 
         & $\mathcal{L}_{SmoothL1}$ & \textcolor{brightgreen}{\textbf{\underline{89.27}}} & 75.22 & 45.37 & 69.24 & 75.54 & 72.89 & 79.29 & 90.89 & 81.03 & 83.26 & 58.82 & \textcolor{brightgreen}{\textbf{\underline{63.13}}} & 63.40 & 62.21 & 37.41 & 69.80 \\ 
         & $\mathcal{L}_{GWD}$ & 88.79 & \textcolor{brightgreen}{\textbf{\underline{77.06}}} & \textcolor{red}{\textbf{49.70}} & \textcolor{red}{\textbf{72.94}} & \textcolor{red}{\textbf{78.08}} & \textcolor{red}{\textbf{77.93}} & \textcolor{brightgreen}{\textbf{\underline{87.45}}} & \textcolor{red}{\textbf{90.90}} & 83.61 & 83.28 & 60.01 & 62.84 & 65.77 & \textcolor{red}{\textbf{66.00}} & \textcolor{red}{\textbf{47.98}} & \textcolor{red}{\textbf{72.82}} \\ 
         & $\mathcal{L}_{KLD}$ & \textcolor{red}{\textbf{89.20}} & 75.56 & 48.32 & \textcolor{brightgreen}{\textbf{\underline{73.02}}} & 76.87 & 75.29 & 86.35 & 90.85 & \textcolor{red}{\textbf{84.53}} & 83.46 & \textcolor{red}{\textbf{60.87}} & 62.13 & 66.55 & 64.90 & 43.86 & 72.12 \\ 
         & $\mathcal{L}_{KFIoU}$ & 89.05 & 75.16 & 49.06 & 69.67 & 78.07 & 75.46 & 86.69 & \textcolor{red}{\textbf{90.90}} & 83.66 & 84.49 & \textcolor{brightgreen}{\textbf{\underline{62.17}}} & \textcolor{red}{\textbf{62.87}} & \textcolor{red}{\textbf{66.72}} & 65.95 & \textcolor{brightgreen}{\textbf{\underline{49.11}}} & 72.60 \\ 
          & $\mathcal{L}_{BD}$ & 88.96 & \textcolor{red}{\textbf{76.84}} & \textcolor{brightgreen}{\textbf{\underline{51.21}}} & 71.75 & \textcolor{brightgreen}{\textbf{\underline{78.56}}} & \textcolor{brightgreen}{\textbf{\underline{79.67}}} & \textcolor{red}{\textbf{87.22}} & \textcolor{brightgreen}{\textbf{\underline{90.91}}} & \textcolor{brightgreen}{\textbf{\underline{85.82}}} & \textcolor{brightgreen}{\textbf{\underline{84.69}}} & 59.15 & 62.73 & \textcolor{brightgreen}{\textbf{\underline{68.08}}} & \textcolor{brightgreen}{\textbf{\underline{67.93}}} & 47.67 & \textcolor{brightgreen}{\textbf{\underline{73.41}}} \\ 
         \hline
    \end{tabular}
    }
    \label{tab:ap_per_category}
\end{table*}

\section{Experiments}
\label{sec:resutls}
\subsection{Dataset}
We conducted our experiments on multiple common datasets for oriented object detection, including DOTA\cite{xia2018dota} and HRSC2016\cite{liu2017high} datasets.

The DOTA\cite{xia2018dota} dataset consists of 2,806 large aerial images from different sensors and platforms. DOTA objects are divided into 15 categories: Plane (PL), Baseball diamond (BD), Bridge (BR), Ground field track (GFT), Small vehicle (SV), Large Vehicle (LV), Ship (SH), Tennis court (TC), Basketball court (BC), Storage tank (ST), Soccer-ball field (SBF), Roundabout (RA), Harbor (HA), Swimming pool (SP), and Helicopter (HC). The training and validation sets contain 1411 and 458 images, respectively; remaining images are used for the testing set. The ground-truth annotations for the testing set are not public; an evaluation server is built for testing. 

The HRSC2016 dataset\cite{liu2017high} is an essential benchmark in high-resolution remote sensing, specifically designed for the detection of maritime vessels in complex environments. Comprising more than 1,000 high-definition images, this dataset offers a comprehensive collection of various ship types, captured under diverse and challenging conditions that mimic real-world scenarios. Each image is accompanied by detailed annotations, which total thousands of precise labels that specify ship locations, orientations, and bounding boxes.

\begin{figure}
    \centering
    \includegraphics[width=1.0\linewidth]{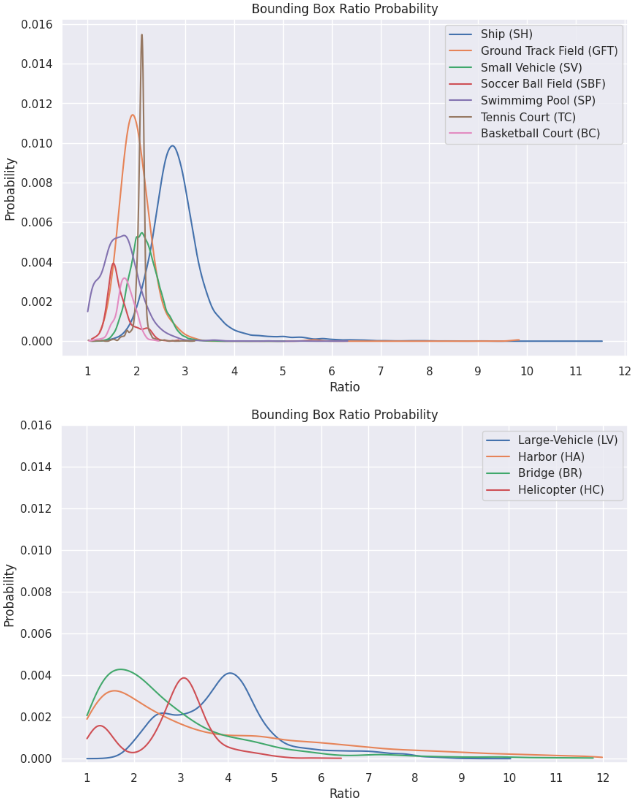}
    \caption{Illustration of the probability distributions of bounding box aspect ratios for different categories of objects, providing insights into their variability. The top plot shows categories with consistent aspect ratios, as indicated by their sharp peaks. In contrast, the bottom plot displays categories with more diverse aspect ratios, evident from their broader and more varied distributions.}
    \label{fig:plot_cls}
\end{figure}
\subsection{Training protocol}
We adopt MMRotate open-source toolbox \cite{zhou2022mmrotate} to conduct our experiments. In all experiments, we ultilize RetinaNet \cite{ross2017focal} and R3Det \cite{yang2021r3det} with the ResNet50 \cite{he2016deep} backbone network architecture for detection frameworks.

In the context of this paper, the model input dimensions for the DOTA-v1.0 dataset are set to 1024$\times$1024 pixels, whereas for the HRSC2016 dataset, the input dimensions are configured to 800$\times$800 pixels. Data preprocessing included normalization and extensive augmentation techniques, including random cropping, randon flipping with ratios of 0.25 for each direction (horizontal, vertical, and diagonal).

Training is conducted over 20 epochs for the DOTA-v1.0 dataset and 50 epochs for the HRSC2016 dataset. The chosen optimizer is AdamW \cite{loshchilov2017decoupled} with an initial learning rate of 1e-4. The learning rate is reduced by the cosine annealing strategy with a minimum value of 1e-8 to ensure stable convergence. We employed a batch size of 2 in all experiments.

\subsection{Experimental Results}
\textbf{Results on DOTA dataset}: Table \ref{tab:ap_per_category} presents the evaluation results of two object detectors, RetinaNet \cite{ross2017focal} and R3Det \cite{yang2021r3det}, on the DOTA-1.0 test set across various object categories. The table highlights the performance metrics for each model using different loss functions, specifically SmoothL1 Loss ($\mathcal{L}_{SmoothL1}$) \cite{girshick2015fast}, Generalized Wasserstein Distance Loss ($\mathcal{L}_{GWD}$) \cite{yang2021rethinking}, Kullback-Leiber Divergence Loss ($\mathcal{L}_{KLD}$) \cite{yang2021learning}, KFIoU Loss ($\mathcal{L}_{KFIoU}$) \cite{yang2022kfiou}, and our proposed Bhattacharyya Distance Loss ($\mathcal{L}_{BD}$) for rotated object detection. The Average Precision (AP) for each class and the overall Average Precision at IoU threshold of 0.50 (AP\(_{50}\)) are provided to quantify the detection performance. The \textbf{\textcolor{brightgreen}{\underline{underlined green}}} and \textbf{\textcolor{red}{red}} results indicate the best and second best performance, respectively. 

\begin{figure*}[htbp]
    \centering
    \includegraphics[width=0.97\linewidth]{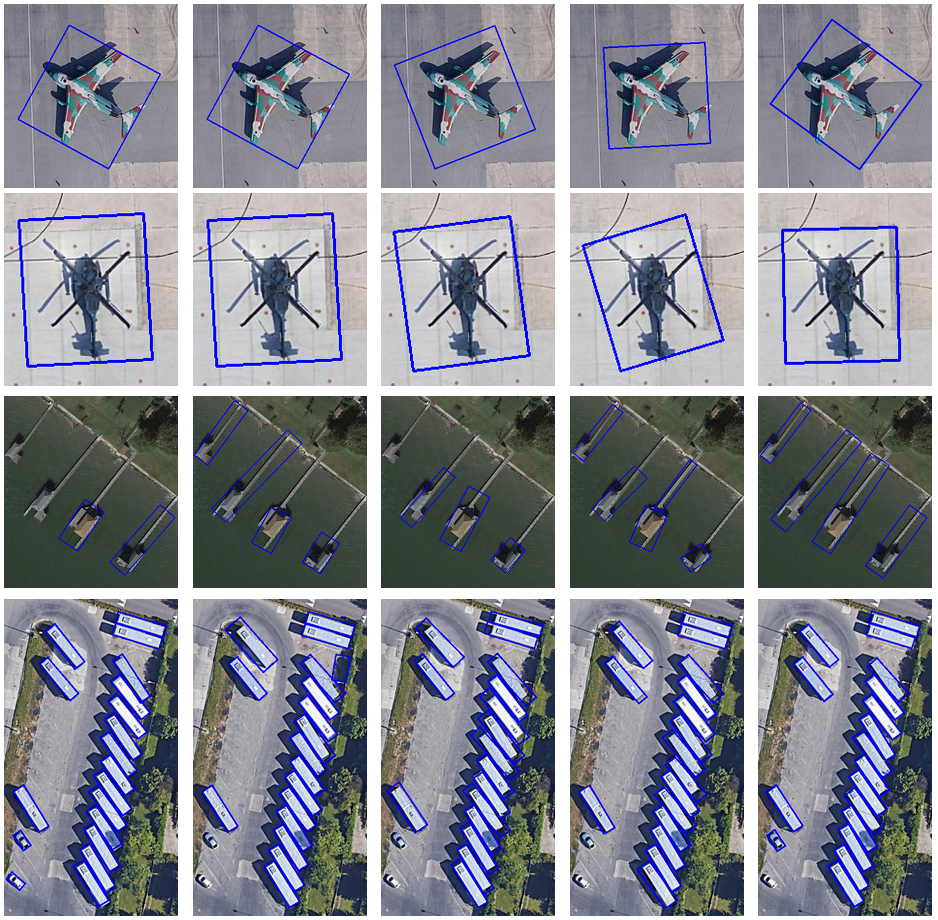}
    \caption{Detection examples using RetinaNet detector with different regression loss functions on DOTA-1.0 test set. From left to right: $\mathcal{L}_{SmoothL1}$, $\mathcal{L}_{GWD}$, $\mathcal{L}_{KLD}$, $\mathcal{L}_{KFIoU}$, and $\mathcal{L}_{BD}$}
    \label{fig:prediction_dota}
\end{figure*}

Focusing on the effect of the Bhattacharyya Distance Loss function, it is evident that it consistently yields the highest (AP\(_{50}\)) scores for both RetinaNet and R3Det object detectors, indicating robust overall performance. For RetinaNet, the Bhattacharyya Distance Loss function achieves an (AP\(_{50}\)) of 71.86\%, which is significantly higher compared to the other loss functions — 69.86\% (+\textbf{\textcolor{brightgreen}{3.43}}\%) for Smooth L1 Loss, 70.07\% (+\textbf{\textcolor{brightgreen}{1.79}}\%) for GWD Loss, 70.31\% (+\textbf{\textcolor{brightgreen}{1.55}}\%) for KLD Loss, and 69.96\% (+\textbf{\textcolor{brightgreen}{1.90}}\%) for KFIoU Loss. This trend is mirrored in the R3Det model, where Bhattacharyya Distance Loss secures the highest (AP\(_{50}\)) score of 73.41\%, outperforming the other loss functions which achieve (AP\(_{50}\)) scores of 69.80\% (+\textbf{\textcolor{brightgreen}{3.61}}\%, Smooth L1 Loss), \textbf{\textcolor{brightgreen}{72.82}} (+0.59\%, GWD Loss), 72.12\% (+\textbf{\textcolor{brightgreen}{1.29}}\%, KLD Loss), and 72.60\% (+\textbf{\textcolor{brightgreen}{0.81}}\%, KFIoU Loss). Analyzing category-specific performance, Bhattacharyya Distance Loss also excels in many individual categories, suggesting that the proposed loss function not only enhances overall performance but also drives improvements in detecting various object categories, demonstrating its effectiveness in object detection tasks. The per-category APs reveal that certain object categories like Plane (PL) and Tennis Court (TC) exhibit consistently high precision across different models and loss functions. In contrast, categories like Bridge (BR) experience lower APs, indicating potential areas for further optimization in model performance. Overall, the evaluation highlights the superiority of the Bhattacharyya Distance Loss in enhancing detection performance for both models, as evidenced by the highest average precision at threshold of 0.50 values. This underscores the importance of selecting appropriate loss functions to achieve optimal detection accuracy across a variety of object categories. With detection examples from the test set of DOTA-v1.0 (illustrated in Figure \ref{fig:prediction_dota}), we can show that the proposed Bhattacharyya Distance loss are able to localize rotated objects more accurately than others on both shape and angle regressions, thus it can detect more true positive objects. 

\ul{By analyzing the results presented in Table} \ref{tab:ap_per_category}, \ul{we observed that serveral categories achieved significantly better performance, including Bridge (BR), Large-Vehicle (LV), Harbor (HA), and Helicopter (HC). To further understand this improvement, we visualized the distribution of class-wise bounding boxes' aspect ratios using Gaussian Kernel Distribution Estimation (as illustrated in Figure} \ref{fig:plot_cls}). \ul{Categories in the top plot demonstrates higher peaks and narrower distributions, indicating consistent aspect ratio. This consistency allows models to predict these bounded shapes with higher accuracy. In contrast, the bottom plot shows categories with broader and more varied distributions, signifying a wider range of aspect ratios. This diversity suggests variability in shapes and orientations, especially for objects with irregular or elongated forms. Our proposed regression loss addresses these challenges by helping detection models avoid overfitting to specific shapes and encouraging the model to generalize better across different object aspect ratios. By leveraging Bhattacharyya distance, which is less sensitive to outliers due to its focus on overall overlap between Gaussian distributions, $\mathcal{L}_{BD}$ loss mitigates the impact of dominant or atypical bounding box shapes on the model’s learning process, fostering balanced attention across all shapes. Furthermore, it is particularly effective in object detection for large aspect ratio bounding boxes. Specifically, the Bhattacharyya Distance loss function assigns greater penalties for mismatches in three critical aspects: the length of the shorter edge, the center point’s position along the shorter edge’s direction, and the angular alignment. These attributes are particularly advantageous, as IoU is inherently sensitive to variations in these aspects when matching bounding boxes with large aspect ratios.}

\textbf{Effectiveness of Anisotropic Gaussian Bounding Box}: Table \ref{tab:mean_ap} presents evaluates the performance of two object detection models, RetinaNet and R3Det, using various loss functions under different bounding box representations (original Gaussian Bounding Box Representation - GBB and Anisotropic Gaussian Bounding Box for square-like objects - AGBB). The performance metrics include average precision at IoU thresholds of 0.5 (AP\(_{50}\)) and 0.75 (AP\(_{75}\)), as well as the mean Average Precision (mAP).

\begin{table}[htbp]
\renewcommand{\arraystretch}{1.25}
    \centering
    \caption{Evaluation of the performance of various loss functions and bounding box representations under Average Precision at different IoU thresholds.}
    \begin{tabular}{|c|c|p{1.5cm}|c|c|c|}
    \hline
        \textbf{Model} & \textbf{Rep.} &\textbf{Loss} & $\textbf{AP}_{50}$ & $\textbf{AP}_{75}$ & \textbf{mAP} \\\hline
        \multirow{5}{*}{\shortstack{RetinaNet\\ \cite{ross2017focal}}} 
         & GBB & $\mathcal{L}_{Smooth L1}$ & 68.43 & 42.02 & 40.12  \\
         & GBB & $\mathcal{L}_{GWD}$ & 70.07 & 41.37 & 41.82 \\
         
         & GBB & $\mathcal{L}_{KLD}$ & 70.31 & 39.49 & 39.73 \\
        
        & GBB & $\mathcal{L}_{KFIoU}$ & 69.96 & 39.60 & 39.74 \\
         & GBB & $\mathcal{L}_{BD}$ & \textbf{71.86} & 42.96 & 42.62 \\ 
         & AGBB & $\mathcal{L}_{BD}$ & 71.05 & \textbf{44.21} & \textbf{42.65} \\ \hline

         \multirow{5}{*}{\shortstack{R3Det\\ \cite{yang2021r3det}}} 
         & GBB &$\mathcal{L}_{SmoothL1}$& 69.80 & 36.58 & 37.81  \\
         & GBB & $\mathcal{L}_{GWD}$ & 72.82 & 39.47 & 40.86 \\
         
         & GBB & $\mathcal{L}_{KLD}$ & 72.12 & 37.10 & 39.54 \\
        
       & GBB  & $\mathcal{L}_{KFIoU}$ & 72.60 & 36.01 & 38.87 \\
         & GBB & $\mathcal{L}_{BD}$ & 73.41 & 42.10 & 42.13 \\ 
         & AGBB & $\mathcal{L}_{BD}$ & \textbf{73.67} & \textbf{43.05} & \textbf{42.81} \\
          \hline    
    \end{tabular}
    \label{tab:mean_ap}
\renewcommand{\arraystretch}{1.25}
\end{table}

\begin{table*}[ht]
\renewcommand{\arraystretch}{1.25}
\caption{Average Precision at different thresholds on HRSC2016 dataset }
    \centering
    \scalebox{1.1}{
    \begin{tabular}{|c|c|c|c|c|c|c|c|c|c|c|c|c|}
        \hline
        \textbf{Model} & \textbf{Loss} & $\textbf{AP}_{50}$ & $\textbf{AP}_{55}$ & $\textbf{AP}_{60}$ & $\textbf{AP}_{65}$ & $\textbf{AP}_{70}$ & $\textbf{AP}_{75}$ & $\textbf{AP}_{80}$ & $\textbf{AP}_{85}$ & $\textbf{AP}_{90}$ & $\textbf{AP}_{95}$ & \textbf{mAP}\\
        \hline
        \multirow{5}{*}{RetinaNet} 
         & $\mathcal{L}_{Smooth L1}$ & 83.3 & 74.7 & 72.3 & 69.6 & 58.1 & 46.2 & 28.0 & 14.4 & 1.70 & 0.10 & 44.82 \\
         & $\mathcal{L}_{GWD}$ & 85.5 & 85.0 & 84.4 & 81.5 & 71.7 & 56.9 & 36.0 & 18.2 & 3.30 & \textcolor{red}{\textbf{0.80}} & 52.33\\
         
         & $\mathcal{L}_{KLD}$ &  \textcolor{brightgreen}{\textbf{\underline{85.8}}} &  \textcolor{brightgreen}{\textbf{\underline{85.5}}} &  \textcolor{brightgreen}{\textbf{\underline{84.8}}} &  \textcolor{brightgreen}{\textbf{\underline{83.1}}} & \textcolor{red}{\textbf{72.5}} & \textcolor{red}{\textbf{61.0}} & \textcolor{red}{\textbf{45.6}} & \textcolor{red}{\textbf{21.2}} &  \textcolor{brightgreen}{\textbf{\underline{8.10}}} & 0.20 & \textcolor{red}{\textbf{54.78}}\\
        
        & $\mathcal{L}_{KFIoU}$ & 85.3 & 84.9 & 83.2 & 74.1 & 68.3 & 48.0 & 28.9 & 14.2 & 4.50 & 0.20 & 49.15 \\
         & $\mathcal{L}_{BD}$ & \textcolor{red}{\textbf{85.7}} & \textcolor{red}{\textbf{85.2}} & \textcolor{red}{\textbf{84.7}} & \textcolor{red}{\textbf{82.8}} & \textcolor{brightgreen}{\textbf{\underline{74.4}}} & \textcolor{brightgreen}{\textbf{\underline{63.3}}} & \textcolor{brightgreen}{\textbf{\underline{48.4}}} & \textcolor{brightgreen}{\textbf{\underline{27.1}}} & \textcolor{red}{\textbf{7.90}} & \textcolor{brightgreen}{\textbf{\underline{3.00}}} & \textcolor{brightgreen}{\textbf{\underline{56.25}}}\\
          
         \hline
         \multirow{5}{*}{R3Det} 
         & $\mathcal{L}_{Smooth L1}$ & 87.9 & 80.9 & 80.5 & 79.5 & 70.1 & 58.8 & 44.9 & 23.0 & 6.60 & \textcolor{brightgreen}{\textbf{\underline{4.50}}} & 53.68\\
         & $\mathcal{L}_{GWD}$ & 89.3 & 88.4 & 80.9 & 80.1 & 70.8 & 66.8 & 47.4 & \textcolor{brightgreen}{\textbf{\underline{26.7}}} & \textcolor{brightgreen}{\textbf{\underline{11.3}}} & \textcolor{red}{\textbf{2.30}} & 56.42\\
         
         & $\mathcal{L}_{KLD}$ & \textcolor{red}{\textbf{89.9}} & \textcolor{red}{\textbf{89.1}} & \textcolor{red}{\textbf{81.1}} & \textcolor{brightgreen}{\textbf{\underline{80.9}}} & \textcolor{brightgreen}{\textbf{\underline{79.5}}} & \textcolor{brightgreen}{\textbf{\underline{68.7}}} & \textcolor{brightgreen}{\textbf{\underline{53.8}}} & \textcolor{red}{\textbf{25.9}} & \textcolor{red}{\textbf{8.20}} & 0.80 & \textcolor{red}{\textbf{57.79}}\\
         & $\mathcal{L}_{KFIoU}$ & 88.9 & 87.8 & \textcolor{red}{\textbf{81.1}} & 80.0 & 70.6 & 67.6 & \textcolor{red}{\textbf{47.9}} & 24.9 & 5.00 & 0.40 & 55.41\\
         & $\mathcal{L}_{BD}$ & \textcolor{brightgreen}{\textbf{\underline{90.2}}} & \textcolor{brightgreen}{\textbf{\underline{89.6}}} & \textcolor{brightgreen}{\textbf{\underline{88.0}}} & \textcolor{red}{\textbf{80.5}} & \textcolor{red}{\textbf{79.3}} & \textcolor{red}{\textbf{67.9}} & 47.1 & 24.4 & 7.10 & \textcolor{brightgreen}{\textbf{\underline{4.50}}} & \textcolor{brightgreen}{\textbf{\underline{57.86}}}\\
         \hline
    \end{tabular}
    }
    \label{tab:hrsc2016}
\renewcommand{\arraystretch}{1.25}
\end{table*}

\begin{figure}
    \centering
    \includegraphics[width=0.9\linewidth]{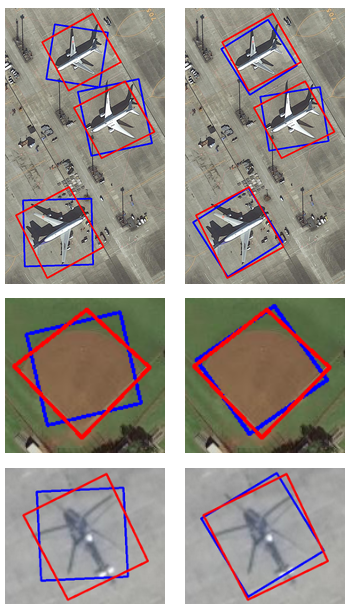}
    \caption{Visual comparison between GBB (left) and AGBB (right) representation on DOTA-v1.0 train/val dataset. The \textbf{\textcolor{red}{red}} and \textbf{\textcolor{blue}{blue}} boxes denote the ground-truth and model predictions, respectively.}
    \label{fig:agbb}
\end{figure}

For the RetinaNet model, the results demonstrate that the Bhattacharyya Distance Loss function performs well across all metrics, with an AP\(_{50}\) of 71.86\%, AP\(_{75}\) of 42.96\%, and mAP of 42.62\% under the Gaussian Bounding Box representation. However, when using the Anisotropic Gaussian representation for square-like bounding boxes with the similar loss function, RetinaNet achieves slightly different results with an AP\(_{50}\) of 71.05\% (-\textbf{\textcolor{red}{0.81}}\%), which is slightly lower than the GBB representation, but shows a notable improvement in the AP\(_{75}\) metric with a value of 44.21\% (+\textcolor{brightgreen}{\textbf{1.25}}\%) and the highest mAP of 42.65\% (+\textcolor{brightgreen}{\textbf{0.03}}\%). This indicates that the AGBB representation for square-like objects enhances the model's performance at higher IoU thresholds, making it more effective for precision tasks. Similarly, for the R3Det model, the results using Bhattacharyya Distance Loss function with GBB representation show strong performance with an AP\(_{50}\) of 73.41\%, AP\(_{75}\) of 42.10\%, and mAP of 42.13\%. However, employing the AGBB representation with the similar loss function significantly boosts the model's AP\(_{50}\) to 73.67\% (+\textcolor{brightgreen}{\textbf{0.26}}\%) and AP\(_{75}\) to 43.05\% (+\textcolor{brightgreen}{\textbf{0.95}}\%), both of which are the highest in the table. The mAP also increases to 42.81 (+\textcolor{brightgreen}{\textbf{0.68}}\%), indicating a better overall performance. The effectiveness of the AGBB representation is evident in the improved scores across both models and multiple metrics. The improvements are more pronounced at the higher IoU threshold (AP\(_{75}\)), suggesting that AGBB representation enhances the precision of detections, particularly in more stringent matching criteria. The higher mAP values further confirm the overall better performance when using AGBB representation compared to the GBB representation under the same loss function. Detection examples shown in Figure \ref{fig:agbb} indicate that AGBB is more effective by providing more accurate and well-aligned square-like bounding boxes compared to GBB representation, improving object detection performance across various scenarios.

In summary, the implementation of the AGBB representation with the  Bhattacharyya Distance loss function demonstrates notable improvements in object detection performance for both RetinaNet and R3Det detectors, particularly at higher precision thresholds, thus proving the AGBB representation's effectiveness in enhancing detection accuracy and overall performance metrics.

\textbf{Results on HRSC2016 dataset}: Table \ref{tab:hrsc2016} presented provides a comprehensive evaluation of Average Precision (AP) metrics at varying Intersection over Union (IoU) thresholds for two object detection models, RetinaNet and R3Det, applied to the HRSC2016 dataset. The evaluation considers a range of IoU thresholds from 50\% to 95\% (denoted as AP\(_{50}\) to AP\(_{95}\)), with the mean Average Precision (mAP) indicating overall performance by averaging AP across all thresholds. The comparison is conducted using different loss functions: Smooth L1 Loss ($\mathcal{L}_{\text{smoothL1}}$), Generalized Wasserstein Distance Loss ($\mathcal{L}_{\text{GWD}}$), Kullback-Leibler Divergence Loss ($\mathcal{L}_{\text{KLD}}$), Kalman Filter IoU Loss ($\mathcal{L}_{\text{KFIoU}}$), and Bhattacharyya Distance Loss ($\mathcal{L}_{\text{BD}}$).

For the RetinaNet model, the Bhattacharyya Distance loss function demonstrates notable effectiveness, achieving the similar AP values as KLD loss at several key thresholds: 85.7\% AP\(_{50}\) (-\textcolor{red}{\textbf{0.1}}\%), 85.2\% AP\(_{55}\) (-\textcolor{red}{\textbf{0.3}}\%), 84.7\% AP\(_{60}\) (-\textcolor{red}{\textbf{0.1}}\%), and 82.8\% AP\(_{65}\) (-\textcolor{red}{\textbf{0.3}}\%). Additionally, it records superior performance at more stringent IoU levels such as 74.4\% AP\(_{70}\) (+\textcolor{brightgreen}{\textbf{1.9}}\%), 63.3\% AP\(_{75}\) (+\textcolor{brightgreen}{\textbf{2.3}}\%), 48.4\% AP\(_{80}\) (+\textcolor{brightgreen}{\textbf{2.8}}\%), 27.1\% AP\(_{85}\) (+\textcolor{brightgreen}{\textbf{5.9}}\%), and 3.00\% AP\(_{95}\) (+\textcolor{brightgreen}{\textbf{2.2}}\%)
while the mAP of 56.25\% underscores its robustness across a spectrum of IoU thresholds. This consistent superiority, especially at higher thresholds, highlights the effectiveness of \(L_{\text{BD}}\) in enhancing the precision of object detection models trained under varied IoU constraints. Similarly, the \(L_{\text{BD}}\) loss function proves to be highly effective for the R3Det model, achieving the highest AP scores at multiple thresholds. The mAP value of 57.86\% further cements its status as a leading performer across all considered loss functions. This superior and consistent performance at both lower and higher IoU thresholds confirms the capability of Bhattacharyya Distance loss in optimally guiding model training to enhance precision and overall detection accuracy. Because ship objects in HRSC2016 dataset have non-square shapes, we do not produce experimental results when training object detection model with AGBB representation on this dataset.

In conclusion, the Bhattacharyya Distance Loss emerges as a highly effective loss function for both the RetinaNet and R3Det models on the HRSC2016 dataset. It consistently achieves or matches the highest AP values across multiple IoU thresholds, indicating its robustness and reliability in fine-tuning object detection models for enhanced precision. This positions our proposed loss function as a superior choice for optimizing detection performance in object detection tasks, warranting further exploration and application in related research and practical implementations.

\section{Conclusion and Future Works}
\label{sec:conclusion}
This paper addresses key challenges in the representation and measurement of overlap in oriented object detection. Recognizing that the original Gaussian distribution is insufficient for square-like objects, we proposed a novel approach that anisotropically scales the Gaussian distribution to better fit these shapes. We further refined our model by applying the Bhattacharyya Distance to compute overlaps between rotated bounding boxes, aligning it with the Intersection over Union (IoU) loss for enhanced accuracy. This innovative approach provides a more precise evaluation of overlap.

Extensive experiments conducted on the DOTA and HRSC2016 datasets demonstrated the robustness of our approach. By integrating the advanced loss function into a state-of-the-art deep learning framework, we observed significant improvements in mean Average Precision metrics, surpassing current methods. Our contributions offer substantial advancements in the field, enhancing the accuracy and reliability of oriented object detection techniques. 

However, experimental results on DOTA dataset show  suboptimal performance of detection frameworks across specific categories within the datasets. Certain object types exhibit lower detection accuracy, highlighting the need for targeted improvements.



\bibliographystyle{cas-model2-names}

\bibliography{cas-refs}





\end{document}